\documentclass[10pt,twocolumn,letterpaper]{article}

\usepackage[pagenumbers]{cvpr} 

\usepackage[dvipsnames]{xcolor}

\usepackage{amssymb}
\usepackage{pifont}

\usepackage{multirow}
\usepackage{adjustbox}
\usepackage{pgfplots}
\pgfplotsset{compat=newest}
\usepackage{colortbl}
\usepackage[accsupp]{axessibility}
\usepackage{siunitx}

\definecolor{cvprblue}{rgb}{0.21,0.49,0.74}
\usepackage[pagebackref,breaklinks,colorlinks,citecolor=cvprblue]{hyperref}

\def\confName{ECCV}
\def\confYear{2024}

\title{First Place Solution to the ECCV 2024 BRAVO Challenge: \\ Evaluating Robustness of Vision Foundation Models for Semantic Segmentation}

\author{Tommie Kerssies, Daan de Geus, Gijs Dubbelman\\
Eindhoven University of Technology \\
{\tt\small \{t.kerssies, d.c.d.geus, g.dubbelman\}@tue.nl}
}

\begin{document}
\maketitle
\begin{abstract}
    In this report, we present the first place solution to the ECCV 2024 BRAVO Challenge, where a model is trained on Cityscapes and its robustness is evaluated on several out-of-distribution datasets. Our solution leverages the powerful representations learned by vision foundation models, by attaching a simple segmentation decoder to DINOv2 and fine-tuning the entire model. This approach outperforms more complex existing approaches, and achieves first place in the challenge. Our code is publicly available\footnote{\url{https://github.com/tue-mps/benchmark-vfm-ss}}.
\end{abstract}

\section{Introduction}
The ECCV 2024 BRAVO Challenge ``aims to benchmark semantic segmentation models on
urban scenes undergoing diverse forms of natural degradation and
realistic-looking synthetic corruption"~\cite{vu2024bravo}. In this report,
we present our solution for Track 1 of this challenge, where a model is trained
on a single labeled semantic segmentation dataset,
Cityscapes~\cite{cordts2016cityscapes}, and evaluated on a range of other
datasets with out-of-distribution image conditions and semantic content, to
evaluate the robustness of the model.

When trained on a single dataset like Cityscapes, vanilla semantic segmentation
models typically only perform well on data that is similar to this training
dataset. Under a distribution shift -- \eg, changing weather conditions,
different geography or image distortion -- the performance drops. To address
this problem, we do not propose a new algorithmic approach or model design, but
instead aim to leverage the power of Vision Foundation Models (VFMs). VFMs are
pre-trained on broad datasets, allowing them to learn powerful representations
for a large variety of downstream
tasks~\cite{oquab2023dinov2,fang2023eva2,radford2021learning,zhai2023sigmoid,bao2021beit,wang2022image}.
As these VFMs learn representations from such diverse data, captured under a
wide array of conditions, we hypothesize that they are perfectly suited for
robust semantic segmentation with simple, vanilla fine-tuning.

In this report, we present our simple meta-approach, and evaluate several
configurations. In our default configuration, we use a
DINOv2~\cite{oquab2023dinov2} VFM, attach a simple linear decoder for
segmentation, and fine-tune the entire model. We assess the impact of using
different model sizes, patch sizes, pre-training strategies and segmentation
decoders. With our simple approach, we significantly outperform existing
specialist models and obtain the first place in the challenge.
Moreover, we make some new observations which could be of interest to future
work.
\begin{figure}[!t]
    \centering
    \includegraphics[width=1\linewidth]{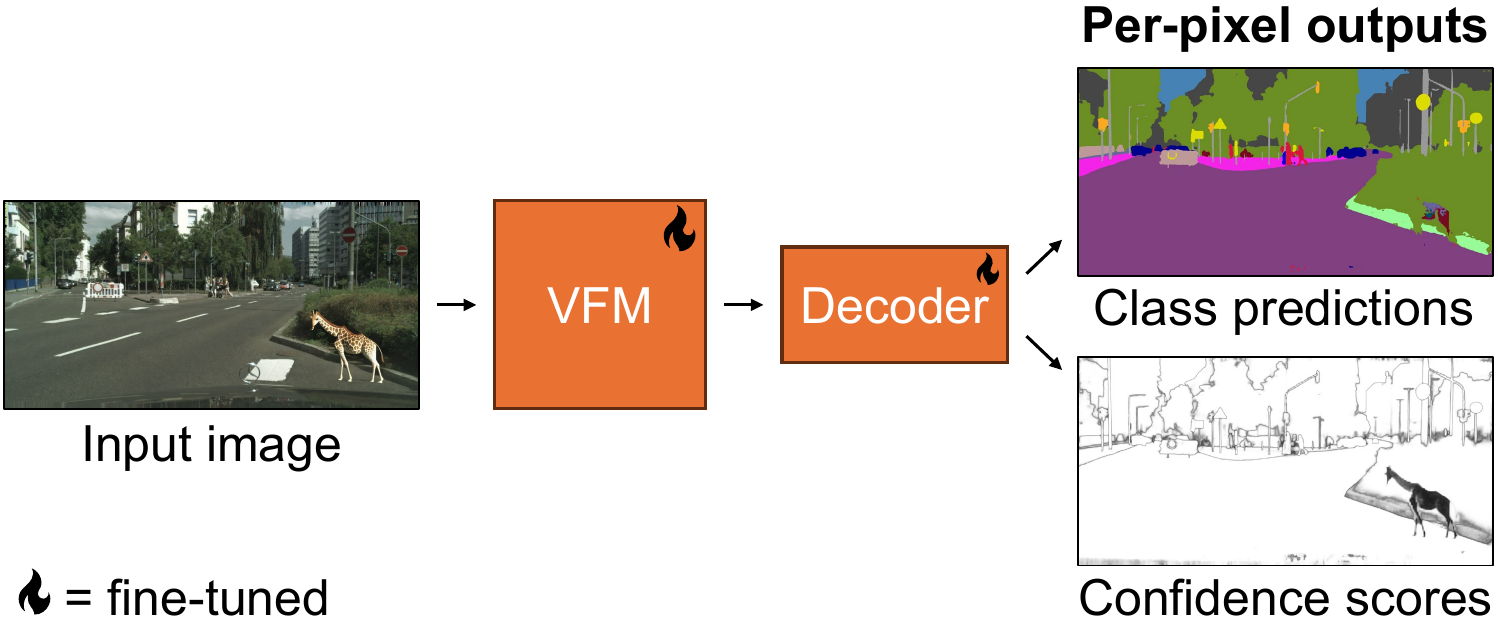}
    \caption{\textbf{Our meta-approach.} We take a pre-trained Vision Foundation Model (VFM), attach a simple segmentation decoder, and fine-tune the entire model for semantic segmentation. The segmentation decoder outputs both the per-pixel classification predictions and the associated confidence scores.}
    \label{fig:arch}
\end{figure}

\section{Method}
The concept of our approach is to fine-tune pre-trained Vision Foundation
Models (VFMs) for semantic segmentation, leveraging the robust representations
these models provide. The meta-approach is visualized in \Cref{fig:arch}. Given
a pre-trained VFM, we attach an off-the-shelf segmentation decoder, and
fine-tune the entire model for semantic segmentation. We evaluate this
meta-architecture in several different configurations.

Our primary solution, which achieves first place in the
challenge, uses the DINOv2 VFM~\cite{oquab2023dinov2}. We selected DINOv2 due
to its demonstrated effectiveness in domain-generalized semantic segmentation
for urban scenes~\cite{kerssies2024benchmarking,englert2024exploring}. DINOv2,
built upon the Vision Transformer (ViT)
architecture~\cite{dosovitskiy2020image}, is pre-trained using self-supervised
learning on a vast, curated dataset. We conduct experiments using all available
sizes of DINOv2.

For our default segmentation decoder, we use a simple linear layer that
transforms the patch-level features $\mathbf{F} \in
    \mathbb{R}^{{E}\times\frac{H}{P}\times\frac{W}{P}}$ into segmentation logits
$\mathbf{L} \in \mathbb{R}^{{C}\times\frac{H}{P}\times\frac{W}{P}}$, where $H$
and $W$ represent the height and width of the input image, $P$ denotes the
patch size, $E$ is the feature dimensionality, and $C$ is the number of classes
in the dataset. We opt for a linear layer because we hypothesize that a more
advanced decoder would provide minimal additional benefit given the already
strong representations learned by the VFM. Additionally, a more advanced
decoder could increase overfitting to the training distribution.

To assess the impact of different design choices of our default configuration,
we compare it with alternative configurations. First, to evaluate the impact of
large-scale pre-training with DINOv2, we make a configuration with a
DeiT-III~\cite{touvron2022deit} ViT pre-trained on
ImageNet-1K~\cite{deng2009imagenet} and fine-tune it on Cityscapes. Second, we
train a model with a Mask2Former decoder~\cite{cheng2022masked} to assess the
impact of a more advanced decoder. Finally, where the default patch size used
is $16 \times 16$, we also experiment with the more computationally expensive
$8 \times 8$ patch size.

\subsection{Training}
When training the model with a linear decoder, we bilinearly upsample the
segmentation logits $\mathbf{L} \in
    \mathbb{R}^{{C}\times\frac{H}{P}\times\frac{W}{P}}$ to $\mathbf{L}' \in
    \mathbb{R}^{{C}\times{H}\times{W}}$, and then apply a categorical cross-entropy
loss to these logits and the semantic segmentation ground truth to fine-tune
the model.

When using Mask2Former, the decoder outputs a set of mask logits $\mathbf{M}
    \in \mathbb{R}^{{N}\times{\frac{H}{P}}\times{\frac{W}{P}}}$ and corresponding
class logits $\mathbf{C} \in \mathbb{R}^{{N}\times{({C}+1})}$, where $N$ is the
number of masks and $\mathbf{C}$ includes an additional ``no-object'' class.
During training, following Mask2Former, these mask and class logits are matched
to the ground truth using bipartite matching. The predicted masks are then
supervised with a cross-entropy loss and a Dice loss, and the predicted classes
are supervised with a categorical cross-entropy loss.

\subsection{Testing}
During inference with the linear decoder, we compute per-pixel class confidence
scores by applying a \texttt{softmax} function to the upsampled class logits
$\mathbf{L}'$. For each pixel, the predicted class is the one with the highest
confidence score, and we also output this confidence score.

During inference with the Mask2Former decoder~\cite{cheng2022masked}, we first
bilinearly upsample the mask logits $\mathbf{M} \in
    \mathbb{R}^{{N}\times{\frac{H}{P}}\times{\frac{W}{P}}}$ to the original
resolution, resulting in $\mathbf{M}' \in \mathbb{R}^{{N}\times{H}\times{W}}$.
We then apply a \texttt{sigmoid} function to $\mathbf{M}'$ to obtain the mask
scores: $\mathbf{P}_\textrm{M} = \texttt{sigmoid}(\mathbf{M}')$.

The class logits $\mathbf{C} \in \mathbb{R}^{{N}\times{({C}+1})}$ are converted
to class scores using a \texttt{softmax} function, excluding the ``no object"
class: $\mathbf{P}_\textrm{C} = \texttt{softmax}(\mathbf{C})[..., :-1]$.

The overall per-pixel class confidence scores $\mathbf{P}' \in \mathbb{R}^{C
        \times H \times W}$ are computed by multiplying the mask scores with the class
scores across all masks. Specifically, for each class $c$ and pixel $(h, w)$,
we have:

\begin{equation}
    \mathbf{P}'_{c,h,w} = \sum_{n=1}^{N} \mathbf{P}_{\textrm{M}_{n,h,w}} \cdot \mathbf{P}_{\textrm{C}_{n,c}}
\end{equation}

For each pixel, the predicted class is the one with the highest value in
$\mathbf{P}'$, and we also output this maximum value as the confidence score.

\section{Experimental Setup}
\subsection{Datasets}
Track 1 of the 2024 BRAVO Challenge focuses on single-domain training,
assessing the robustness of models trained with limited supervision and
geographical diversity against real-world distribution shifts. Only the
Cityscapes dataset~\cite{cordts2016cityscapes} is permitted for training. This
dataset features homogeneous urban scenes with fine-grained annotations for 19
semantic classes.

The BRAVO benchmark dataset, used for evaluation, consists of six subsets:
\begin{itemize}
      \item \textbf{ACDC~\cite{SDV21}}: Real scenes captured in adverse weather conditions, such as fog, rain, night, and snow.

      \item \textbf{SMIYC~\cite{BENCHMARKS2021_d67d8ab4}}: Real scenes featuring out-of-distribution (OOD) objects that are rarely encountered on the road.

      \item \textbf{Out-of-context~\cite{franchi2021robust}}: Augmented scenes with random backgrounds, generated by replacing the backgrounds of 329 validation images from Cityscapes with random ones.

      \item \textbf{Synflare~\cite{wu2021train}}: Augmented scenes with synthesized light flares, produced by adding random light flares to 308 validation images from Cityscapes.

      \item \textbf{Synobjs~\cite{loiseau2024reliability}}: Augmented scenes with inpainted synthetic OOD objects, created by adding 26 different OOD objects to 656 validation images from Cityscapes.

      \item \textbf{Synrain~\cite{pizzati2023physics}}: Augmented scenes with synthesized raindrops on the camera lens, generated by augmenting 500 validation images from Cityscapes.
\end{itemize}

\subsection{Evaluation metrics}
The ECCV 2024 BRAVO Challenge evaluates methods based on a variety of metrics to
assess their performance in semantic segmentation and out-of-distribution (OOD)
detection. The metrics are grouped into three categories: semantic metrics, OOD
metrics, and summary metrics. The semantic metrics evaluate the overall quality
of the semantic segmentation predictions, while the OOD metrics assess the
model's ability to detect OOD objects. The summary metrics combine the semantic
and OOD metrics to provide an overall ranking of the methods.

\paragraph{Semantic metrics.} The semantic metrics are computed on all subsets, except SMIYC, for valid
pixels only. Valid pixels are those not “invalidated” by extreme uncertainty,
such as pixels obscured by the brightest areas of a flare or covered by an OOD
object. The semantic metrics include:
\begin{itemize}
      \item \textbf{Mean Intersection over Union (mIoU)}: Measures the proportion of correctly labeled pixels among all pixels. It is the only semantic metric that does not rely on prediction confidence. Higher mIoU values indicate better segmentation accuracy.
      \item \textbf{Expected Calibration Error (ECE)}: Quantifies the difference between predicted confidence and actual accuracy. Lower ECE values indicate better calibration of the model’s confidence scores.
      \item \textbf{Area Under the ROC Curve (AUROC)}: Represents the area under the curve plotting the true positive rate against the false positive rate, using the predicted confidence level to rank pixels. Higher AUROC values indicate better ability to distinguish between correct and incorrect predictions.
      \item \textbf{False Positive Rate at 95\% True Positive Rate (FPR@95)}: Measures the false positive rate when the true positive rate is 95\% computed in the ROC curve above. Lower FPR@95 values indicate better robustness against false positives.
      \item \textbf{Area Under the Precision-Recall Curve (AUPR)}: Represents the area under the curve plotting precision against recall, using the predicted confidence level to rank pixels. Higher AUPR-Success and AUPR-Error values indicate better ability to identify correct and incorrect predictions, respectively.
\end{itemize}

\paragraph{OOD metrics.} The OOD metrics are computed on the SMIYC and Synobjs subsets only. Invalid pixels in these datasets are those that are obscured by
OOD objects. The OOD metrics quantify whether the model attributes, as expected, less confidence to the invalid pixels. The OOD metrics include:
\begin{itemize}
      \item \textbf{Area Under the Precision-Recall Curve (AUPRC)}: AUPRC, over the binary criterion
            of a pixel being invalid, ranked by the reversed predicted confidence level for the
            pixel.

      \item \textbf{Area Under the ROC Curve (AUROC)}: Area Under the ROC Curve, over the binary
            criterion of a pixel being invalid, ranked by the reversed predicted confidence level
            for the pixel.

      \item \textbf{False Positive Rate at 95\% True Positive Rate (FPR@95)}: False Positive Rate
            when True Positive Rate is 95\% computed in the ROC curve above.
\end{itemize}

\paragraph{Summary metrics.} The summary metrics include:
\begin{itemize}
      \item \textbf{Semantic}: The harmonic mean of all semantic metrics, where ECE and FPR@95 are
            reversed.
      \item \textbf{OOD}: The harmonic mean of all OOD metrics, where FPR@95 is reversed.
      \item \textbf{BRAVO}: The harmonic mean of the semantic and OOD harmonic means, serving as the official ranking metric for the challenge.
\end{itemize}

\subsection{Implementation Details}
We use the following models from the \texttt{timm} library~\cite{rw2019timm} to
initialize the VFM:
\begin{itemize}
      \item \texttt{deit3\_small\_patch16\_224.fb\_in1k};
      \item \texttt{vit\_small\_patch14\_dinov2};
      \item \texttt{vit\_base\_patch14\_dinov2};
      \item \texttt{vit\_large\_patch14\_dinov2};
      \item \texttt{vit\_giant\_patch14\_dinov2}.
\end{itemize}

The models are fine-tuned for 40 epochs using two A6000 GPUs, with a batch size
of 1 per GPU and gradient accumulation over 8 steps, resulting in an effective
batch size of 16. Our implementation follows the details provided
in~\cite{kerssies2024benchmarking}. Notably, the learning rate for the VFM
weights is set to be 10$\times$ smaller than the overall learning rate, as this
configuration empirically yields better results. For the Mask2Former decoder,
we employ a variant specifically adapted for use with a single-scale ViT
encoder, as introduced in~\cite{kerssies2024benchmarking}.

\section{Results}

\begin{table*}\footnotesize
    \centering
    \begin{tabular}{lccc}
        \toprule
        Method                                                      & BRAVO $\uparrow$ & Semantic $\uparrow$ & OOD $\uparrow$   \\
        \midrule
        DINOv2, ViT-L, 8x8 patch size, linear decoder               & \textbf{77.9}    & 69.8                & \underline{88.1} \\
        DINOv2, ViT-L, 16x16 patch size, linear decoder             & \underline{77.2} & \textbf{70.8}       & 84.8             \\
        DINOv2, ViT-g, 16x16 patch size, linear decoder             & 76.1             & 70.0                & 83.4             \\
        DINOv2, ViT-B, 16x16 patch size, linear decoder             & 75.5             & \underline{70.5}    & 81.4             \\
        DINOv2, ViT-S, 16x16 patch size, linear decoder             & 69.9             & 69.1                & 70.6             \\
        \textcolor{gray}{PixOOD YOLO (="Model Selection")}          & 67.8             & 57.1                & 83.5             \\
        DINOv2, ViT-g, 16x16 patch size, Mask2Former decoder        & 64.5             & 49.7                & \textbf{92.1}    \\
        \textcolor{gray}{Model selection}                           & 63.5             & 69.4                & 58.5             \\
        \textcolor{gray}{PixOOD w/ ResNet-101 DeepLab}              & 61.2             & 58.7                & 64.0             \\
        \textcolor{gray}{Ensemble C}                                & 61.1             & 64.3                & 58.2             \\
        \textcolor{gray}{Ensemble A}                                & 59.9             & 67.3                & 53.9             \\
        \textcolor{gray}{PixOOD w/ DeepLab Decoder}                 & 59.4             & 46.1                & 83.5             \\
        DeiT III (IN1K), ViT-S, 16x16 patch size, linear decoder    & 54.1             & 62.8                & 47.6             \\
        \textcolor{gray}{PixOOD}                                    & 53.5             & 40.4                & 79.1             \\
        \textcolor{gray}{SegFormer-B5}                              & 47.1             & 45.3                & 49.2             \\
        \textcolor{gray}{ObsNet-R101-DLv3plus}                      & 45.3             & 51.5                & 40.5             \\
        \textcolor{gray}{Mask2Former-SwinB}                         & 37.7             & 27.7                & 59.2             \\
        \textcolor{gray}{Physically Feasible Semantic Segmentation} & 33.6             & 66.3                & 22.5             \\
        \bottomrule
    \end{tabular}
    \caption{\textbf{BRAVO index}. The official ranking metric, calculated as the harmonic mean of the semantic and OOD harmonic means for each method. Grayed-out methods are submitted by other participants.}
    \label{tab:bravo}
\end{table*}

\quad\textbf{BRAVO index.} The official ranking metric, calculated as the harmonic mean of the semantic
and OOD harmonic means for each method, is shown in Table~\ref{tab:bravo}. Our
best-performing model, DINOv2 with a ViT-L/8 backbone and a linear decoder,
achieves the highest BRAVO index of 77.9, which is +10.1 higher than the next best method not submitted by us, which uses multiple different specialist models, \ie, \textit{PixOOD YOLO (="Model Selection")}.

\begin{table*}\footnotesize
    \centering
    \begin{tabular}{lcccccc}
        \toprule
        Method                                                      & ACDC $\uparrow$  & SMIYC $\uparrow$ & Out-of-context $\uparrow$ & Synflare $\uparrow$ & Synobjs $\uparrow$ & Synrain $\uparrow$ \\
        \midrule
        DINOv2, ViT-L, 8x8 patch size, linear decoder               & 67.3             & \underline{89.9} & 71.0                      & 72.7                & \textbf{76.7}      & 73.9               \\
        DINOv2, ViT-L, 16x16 patch size, linear decoder             & \textbf{69.4}    & 89.3             & 70.4                      & 72.4                & \underline{75.1}   & \underline{73.8}   \\
        DINOv2, ViT-g, 16x16 patch size, linear decoder             & 67.7             & 88.2             & 71.0                      & 73.2                & 74.7               & 73.2               \\
        DINOv2, ViT-B, 16x16 patch size, linear decoder             & \underline{68.5} & 87.9             & \underline{71.2}          & 72.8                & 74.0               & 73.0               \\
        DINOv2, ViT-S, 16x16 patch size, linear decoder             & 66.9             & 83.1             & 70.2                      & 70.6                & 68.6               & 72.9               \\
        \textcolor{gray}{PixOOD YOLO (="Model Selection")}          & 55.7             & 74.9             & 64.9                      & 65.0                & 56.4               & 63.3               \\
        DINOv2, ViT-g, 16x16 patch size, Mask2Former decoder        & 49.1             & \textbf{94.4}    & 40.9                      & 53.9                & 64.3               & 60.1               \\
        \textcolor{gray}{Model selection}                           & 66.6             & 70.7             & \textbf{73.4}             & \textbf{77.7}       & 58.9               & \textbf{76.4}      \\
        \textcolor{gray}{PixOOD w/ ResNet-101 DeepLab}              & 55.7             & 53.2             & 64.9                      & 65.0                & 63.2               & 63.3               \\
        \textcolor{gray}{Ensemble C}                                & 64.8             & 63.1             & 61.7                      & 69.6                & 58.1               & 64.6               \\
        \textcolor{gray}{Ensemble A}                                & 66.4             & 58.7             & 66.0                      & \underline{74.0}    & 58.9               & 69.4               \\
        \textcolor{gray}{PixOOD w/ DeepLab Decoder}                 & 48.4             & 74.9             & 56.2                      & 43.6                & 56.4               & 36.4               \\
        DeiT III (IN1K), ViT-S, 16x16 patch size, linear decoder    & 58.8             & 50.1             & 65.1                      & 71.4                & 58.5               & 65.6               \\
        \textcolor{gray}{PixOOD}                                    & 40.8             & 66.7             & 50.0                      & 48.4                & 49.4               & 36.4               \\
        \textcolor{gray}{SegFormer-B5}                              & 45.5             & 60.5             & 39.1                      & 50.6                & 41.7               & 51.5               \\
        \textcolor{gray}{ObsNet-R101-DLv3plus}                      & 50.8             & 38.7             & 50.7                      & 50.9                & 50.2               & 52.0               \\
        \textcolor{gray}{Mask2Former-SwinB}                         & 28.6             & 64.5             & 23.5                      & 41.1                & 29.3               & 26.7               \\
        \textcolor{gray}{Physically Feasible Semantic Segmentation} & 63.6             & 20.9             & 68.5                      & 68.1                & 43.1               & 71.0               \\
        \bottomrule
    \end{tabular}
    \caption{\textbf{Subset harmonic means}. Harmonic means of semantic and OOD metrics for each subset in the BRAVO benchmark dataset, computed for each method. Grayed-out methods are submitted by other participants.}
    \label{tab:subsets}
\end{table*}

\textbf{Subset harmonic means.} The harmonic means of semantic and OOD metrics for each subset in the BRAVO
benchmark dataset are shown in Table~\ref{tab:subsets}. Although our
DINOv2-based models perform well across most subsets, they do not consistently
outperform all other methods. Specifically, on the Out-of-context, Synobjs, and
Synrain subsets, the method named \textit{Model selection} outperforms our
models. Interestingly, compared to other methods, our DINOv2-based models work
particularly well on the datasets with OOD objects; SMIYC and Synobjs. On SMIYC, our
best models achieve scores above 89.0, while the next best method scores 74.9, and
others score significantly lower. This indicates that the VFM pre-training is
particularly helpful for out-of-distribution detection in real-world scenarios.

\begin{table*}\footnotesize
    \centering
    \begin{tabular}{lcccccc}
        \toprule
        Method                                                      & mIoU $\uparrow$  & AUPR-Error $\uparrow$ & AUPR-Success $\uparrow$ & AUROC $\uparrow$ & ECE $\downarrow$ & FPR@95 $\downarrow$ \\
        \midrule
        DINOv2, ViT-L, 8x8 patch size, linear decoder               & 76.7             & 40.0                  & 99.4                    & 91.4             & 2.0              & 38.8                \\
        DINOv2, ViT-L, 16x16 patch size, linear decoder             & 75.9             & 41.2                  & \underline{99.5}        & 92.3             & \underline{1.7}     & 37.8                \\
        DINOv2, ViT-g, 16x16 patch size, linear decoder             & \underline{77.6} & 39.3                  & \textbf{99.5}           & 92.3             & 1.8  & 37.6                \\
        DINOv2, ViT-B, 16x16 patch size, linear decoder             & 71.7             & 43.3                  & 99.4                    & 92.3             & 2.1              & 40.3                \\
        DINOv2, ViT-S, 16x16 patch size, linear decoder             & 66.5             & 45.1                  & 99.2                    & 91.8             & 2.5              & 44.3                \\
        \textcolor{gray}{PixOOD YOLO (Model Selection)}             & 43.6             & \underline{57.3}      & 93.5                    & 83.5             & 15.1             & 55.8                \\
        DINOv2, ViT-g, 16x16 patch size, Mask2Former decoder        & \textbf{78.2}    & 23.2                  & 99.2                    & 87.9             & 5.0              & 63.6                \\
        \textcolor{gray}{Model selection}                           & 73.6             & 44.8                  & 99.1                    & 91.8             & 9.2             & 40.2                \\
        \textcolor{gray}{PixOOD w/ ResNet-101 DeepLab}              & 43.2             & \textbf{58.5}         & 93.5                    & 84.0             & 15.1             & 54.6                \\
        \textcolor{gray}{Ensemble C}                                & 73.9             & 47.4                  & 99.1                    & \underline{92.5} & 52.7             & \textbf{34.7}       \\
        \textcolor{gray}{Ensemble A}                                & 73.8             & 47.0                  & 99.1                    & \textbf{92.6}    & 38.6             & \underline{36.9}    \\
        \textcolor{gray}{PixOOD w/ DeepLab Decoder}                 & 69.3             & 23.5                  & 97.4                    & 76.0             & 13.8             & 69.9                \\
        DeiT III (IN1K), ViT-S, 16x16 patch size, linear decoder    & 44.9             & 54.7                  & 97.9                    & 89.2             & \textbf{1.7}              & 54.2                \\
        \textcolor{gray}{PixOOD}                                    & 64.8             & 22.4                  & 96.3                    & 73.1             & 17.6             & 76.6                \\
        \textcolor{gray}{SegFormer-B5}                              & 67.4             & 24.1                  & 97.2                    & 77.2             & 30.7             & 71.9                \\
        \textcolor{gray}{ObsNet-R101-DLv3plus}                      & 65.3             & 32.1                  & 98.5                    & 87.8             & 45.6             & 63.4                \\
        \textcolor{gray}{Mask2Former-SwinB}                         & 67.2             & 13.2                  & 90.4                    & 47.0             & 55.1             & 82.8                \\
        \textcolor{gray}{Physically Feasible Semantic Segmentation} & 67.4             & 42.2                  & 98.7                    & 89.4             & 4.8              & 48.3                \\
        \bottomrule
    \end{tabular}
    \caption{\textbf{Semantic metrics}. Performance metrics for valid pixel predictions and their confidence, averaged across all subsets except SMIYC, computed for each method. Grayed-out methods are submitted by other participants.}
    \label{tab:semantic}
\end{table*}

\textbf{Semantic metrics.} Table~\ref{tab:semantic} presents the performance metrics for valid pixel
predictions and their confidence, averaged across all subsets except SMIYC.

We observe that the model with the highest mIoU, DINOv2 with a ViT-g/16
backbone and a Mask2Former decoder, performs relatively poorly on the other
metrics. As the other metrics take into account the confidence score, this
suggests that while Mask2Former is effective at predicting the correct class,
it is less adept at estimating the confidence of its predictions, at least in
the out-of-the-box manner in which we used it. In our setup, models with a
simple linear decoder provide the best trade-off between segmentation accuracy
and confidence estimation.

A similar result is observed when changing the patch size. A smaller patch size
of $8 \times 8$ results in better mIoU, but the other metrics are worse. This
indicates that a smaller patch size allows the model to capture more
fine-grained details, which improves the accuracy of the predicted class
labels, but that this somehow makes the confidence scores less reliable.

Another noteworthy observation is that all our models have relatively low ECE
values, indicating that they are well-calibrated, even though no explicit
calibration techniques were applied. Even the DeiT-III-based model, which
scores low on the overall BRAVO score, achieves a low ECE value of 1.7.
Therefore, further investigation is needed to understand why the ECE values are so low.

Finally, the results suggest that more accurate models in terms of mIoU tend to
be worse at identifying their own errors, as indicated by the AUPR-Error
metric. However, they excel at identifying correct predictions, as shown by the
AUPR-Success metric. It is possible that this happens simply because errors by
accurate models are rarer, making it harder to identify them.

Overall, the results show that mIoU, which does not depend on prediction
confidence, does not correlate well with the other metrics that do.

\begin{table*}\footnotesize
    \centering
    \begin{tabular}{lccc}
        \toprule
        Method                                                      & AUPRC $\uparrow$ & AUROC $\uparrow$ & FPR@95 $\downarrow$ \\
        \midrule
        DINOv2, ViT-L, 8x8 patch size, linear decoder               & 81.7             & \underline{97.7} & \underline{12.9}    \\
        DINOv2, ViT-L, 16x16 patch size, linear decoder             & 76.7             & 97.1             & 15.0                \\
        DINOv2, ViT-g, 16x16 patch size, linear decoder             & 74.3             & 96.9             & 15.3                \\
        DINOv2, ViT-B, 16x16 patch size, linear decoder             & 70.6             & 96.6             & 15.1                \\
        DINOv2, ViT-S, 16x16 patch size, linear decoder             & 58.9             & 94.9             & 20.2                \\
        \textcolor{gray}{PixOOD YOLO (Model Selection)}             & 79.0             & 96.5             & 18.2                \\
        DINOv2, ViT-g, 16x16 patch size, Mask2Former decoder        & \textbf{84.1}    & \textbf{98.8}    & \textbf{4.5}        \\
        \textcolor{gray}{Model selection}                           & 52.7             & 93.7             & 26.9                \\
        \textcolor{gray}{PixOOD w/ ResNet-101 DeepLab}              & 47.1             & 91.7             & 20.8                \\
        \textcolor{gray}{Ensemble C}                                & 36.0             & 92.8             & 14.9                \\
        \textcolor{gray}{Ensemble A}                                & 32.3             & 91.4             & 17.2                \\
        \textcolor{gray}{PixOOD w/ DeepLab Decoder}                 & 79.0             & 96.5             & 18.2                \\
        DeiT III (IN1K), ViT-S, 16x16 patch size, linear decoder    & 30.0             & 86.5             & 38.6                \\
        \textcolor{gray}{PixOOD}                                    & \underline{83.0} & 95.2             & 25.7                \\
        \textcolor{gray}{SegFormer-B5}                              & 51.0             & 86.4             & 63.1                \\
        \textcolor{gray}{ObsNet-R101-DLv3plus}                      & 28.4             & 80.8             & 56.3                \\
        \textcolor{gray}{Mask2Former-SwinB}                         & 75.5             & 91.9             & 61.7                \\
        \textcolor{gray}{Physically Feasible Semantic Segmentation} & 26.9             & 83.5             & 74.5                \\
        \bottomrule
    \end{tabular}
    \caption{\textbf{OOD metrics}. Performance metrics for detecting OOD objects by identifying invalid pixels based on prediction confidence, averaged over the SMIYC and Synobjs subsets, computed for each method. Grayed-out methods are submitted by other participants.}
    \label{tab:ood}
\end{table*}

\textbf{OOD metrics.} Table~\ref{tab:ood} presents the performance metrics for detecting OOD objects
by identifying invalid pixels based on prediction confidence, averaged over the
SMIYC and Synobjs subsets.

Surprisingly, our configuration with worst confidence estimation for valid
pixels (see Table~\ref{tab:semantic}), DINOv2 with ViT-g/16 and Mask2Former,
achieves the highest AUPRC of 84.1, the highest AUROC of 98.8, and the lowest
FPR@95 of 4.5 for detecting invalid pixels. This suggests that the mask
classification framework used by Mask2Former, where per-class masks are
predicted separately, allows this decoder to more accurately identify which
pixels belong to the mask of an in-distribution class and which do not.

Additionally, while a smaller patch size results in worse confidence estimation
for valid pixels (see Table~\ref{tab:semantic}), it helps in identifying
invalid pixels. Qualitative analyses show that the smaller patch size enables
the model to better separate valid and invalid pixels, as it can capture more
fine-grained details.

Finally, while scaling from ViT-L to ViT-g improves mIoU for valid pixels (see Table~\ref{tab:semantic}), OOD detection performance shows a noticeable degradation.

Overall, the results indicate that the models best at identifying invalid
pixels are not necessarily the same ones that excel at correctly classifying
valid pixels or accurately estimating their confidence for valid pixels.
\section{Conclusion}
This report presents the winning solution for Track 1 of the 2024 BRAVO
Challenge. Our approach is based on fine-tuning Vision Foundation Models (VFMs)
for semantic segmentation, leveraging the strong representations learned during
pre-training. The results demonstrate that, just through better pre-training, a
fine-tuned VFM is more robust under distribution shifts than complex specialist
models, and can both accurately classify pixels and reliably estimate its
confidence in these predictions. We evaluated our approach in several
configurations, which has lead to several new observations. While we identify
potential causes of these observations, future work is necessary to explore
this further. Particularly, we believe it would be interesting to investigate the reason for the difference in semantic and OOD metrics with the Mask2Former
decoder compared to the linear decoder. Additionally, future works could
explore whether some of the more specialized out-of-distribution detection or
calibration methods are still effective when used in combination with VFMs.

{\small
    \paragraph{Acknowledgements.} This work was supported by Chips Joint Undertaking
    (Chips JU) in EdgeAI “Edge AI Technologies for Optimised Performance Embedded
    Processing” project, grant agreement No 101097300.}

{
    \small
    \bibliographystyle{ieeenat_fullname}
    \bibliography{main}

\begin{thebibliography}{21}
\providecommand{\natexlab}[1]{#1}
\providecommand{\url}[1]{\texttt{#1}}
\expandafter\ifx\csname urlstyle\endcsname\relax
  \providecommand{\doi}[1]{doi: #1}\else
  \providecommand{\doi}{doi: \begingroup \urlstyle{rm}\Url}\fi

\bibitem[Bao et~al.(2022)Bao, Dong, Piao, and Wei]{bao2021beit}
Hangbo Bao, Li Dong, Songhao Piao, and Furu Wei.
\newblock {BEiT: BERT Pre-Training of Image Transformers}.
\newblock In \emph{ICLR}, 2022.

\bibitem[Chan et~al.(2021)Chan, Lis, Uhlemeyer, Blum, Honari, Siegwart, Fua, Salzmann, and Rottmann]{BENCHMARKS2021_d67d8ab4}
Robin Chan, Krzysztof Lis, Svenja Uhlemeyer, Hermann Blum, Sina Honari, Roland Siegwart, Pascal Fua, Mathieu Salzmann, and Matthias Rottmann.
\newblock Segmentmeifyoucan: A benchmark for anomaly segmentation.
\newblock In \emph{NeurIPS}, 2021.

\bibitem[Cheng et~al.(2022)Cheng, Misra, Schwing, Kirillov, and Girdhar]{cheng2022masked}
Bowen Cheng, Ishan Misra, Alexander~G Schwing, Alexander Kirillov, and Rohit Girdhar.
\newblock {Masked-attention Mask Transformer for Universal Image Segmentation}.
\newblock In \emph{CVPR}, 2022.

\bibitem[Cordts et~al.(2016)Cordts, Omran, Ramos, Rehfeld, Enzweiler, Benenson, Franke, Roth, and Schiele]{cordts2016cityscapes}
Marius Cordts, Mohamed Omran, Sebastian Ramos, Timo Rehfeld, Markus Enzweiler, Rodrigo Benenson, Uwe Franke, Stefan Roth, and Bernt Schiele.
\newblock {The Cityscapes Dataset for Semantic Urban Scene Understanding}.
\newblock In \emph{CVPR}, 2016.

\bibitem[Deng et~al.(2009)Deng, Dong, Socher, Li, Li, and Fei-Fei]{deng2009imagenet}
Jia Deng, Wei Dong, Richard Socher, Li-Jia Li, Kai Li, and Li Fei-Fei.
\newblock {ImageNet: A Large-Scale Hierarchical Image Database}.
\newblock In \emph{CVPR}, 2009.

\bibitem[Dosovitskiy et~al.(2021)Dosovitskiy, Beyer, Kolesnikov, Weissenborn, Zhai, Unterthiner, Dehghani, Minderer, Heigold, Gelly, et~al.]{dosovitskiy2020image}
Alexey Dosovitskiy, Lucas Beyer, Alexander Kolesnikov, Dirk Weissenborn, Xiaohua Zhai, Thomas Unterthiner, Mostafa Dehghani, Matthias Minderer, Georg Heigold, Sylvain Gelly, et~al.
\newblock {An Image is Worth 16x16 Words: Transformers for Image Recognition at Scale}.
\newblock In \emph{ICLR}, 2021.

\bibitem[{Englert, Brunó B.} et~al.(2024){Englert, Brunó B.}, {Piva, Fabrizio J.}, {Kerssies, Tommie}, {de Geus, Daan}, and {Dubbelman, Gijs}]{englert2024exploring}
{Englert, Brunó B.}, {Piva, Fabrizio J.}, {Kerssies, Tommie}, {de Geus, Daan}, and {Dubbelman, Gijs}.
\newblock {Exploring the Benefits of Vision Foundation Models for Unsupervised Domain Adaptation}.
\newblock In \emph{CVPRW}, 2024.

\bibitem[Fang et~al.(2023)Fang, Sun, Wang, Huang, Wang, and Cao]{fang2023eva2}
Yuxin Fang, Quan Sun, Xinggang Wang, Tiejun Huang, Xinlong Wang, and Yue Cao.
\newblock {EVA-02: A Visual Representation for Neon Genesis}.
\newblock \emph{arXiv preprint arXiv:2303.11331}, 2023.

\bibitem[Franchi et~al.(2021)Franchi, Belkhir, Ha, Hu, Bursuc, Blanz, and Yao]{franchi2021robust}
Gianni Franchi, Nacim Belkhir, Mai~Lan Ha, Yufei Hu, Andrei Bursuc, Volker Blanz, and Angela Yao.
\newblock Robust semantic segmentation with superpixel-mix.
\newblock In \emph{BMVC}, 2021.

\bibitem[Kerssies et~al.(2024)Kerssies, de~Geus, and Dubbelman]{kerssies2024benchmarking}
Tommie Kerssies, Daan de Geus, and Gijs Dubbelman.
\newblock {How to Benchmark Vision Foundation Models for Semantic Segmentation?}
\newblock In \emph{CVPRW}, 2024.

\bibitem[Loiseau et~al.(2024)Loiseau, Vu, Chen, P{\'e}rez, and Cord]{loiseau2024reliability}
Thibaut Loiseau, Tuan-Hung Vu, Mickael Chen, Patrick P{\'e}rez, and Matthieu Cord.
\newblock Reliability in semantic segmentation: Can we use synthetic data?
\newblock In \emph{ECCV}, 2024.

\bibitem[Oquab et~al.(2023)Oquab, Darcet, Moutakanni, Vo, Szafraniec, Khalidov, Fernandez, Haziza, Massa, El-Nouby, et~al.]{oquab2023dinov2}
Maxime Oquab, Timoth{\'e}e Darcet, Th{\'e}o Moutakanni, Huy Vo, Marc Szafraniec, Vasil Khalidov, Pierre Fernandez, Daniel Haziza, Francisco Massa, Alaaeldin El-Nouby, et~al.
\newblock {DINOv2: Learning Robust Visual Features without Supervision}.
\newblock In \emph{TMLR}, 2023.

\bibitem[Pizzati et~al.(2023)Pizzati, Cerri, and de~Charette]{pizzati2023physics}
Fabio Pizzati, Pietro Cerri, and Raoul de Charette.
\newblock Physics-informed guided disentanglement in generative networks.
\newblock \emph{T-PAMI}, 2023.

\bibitem[Radford et~al.(2021)Radford, Kim, Hallacy, Ramesh, Goh, Agarwal, Sastry, Askell, Mishkin, Clark, et~al.]{radford2021learning}
Alec Radford, Jong~Wook Kim, Chris Hallacy, Aditya Ramesh, Gabriel Goh, Sandhini Agarwal, Girish Sastry, Amanda Askell, Pamela Mishkin, Jack Clark, et~al.
\newblock {Learning Transferable Visual Models From Natural Language Supervision}.
\newblock In \emph{ICML}, 2021.

\bibitem[Sakaridis et~al.(2021)Sakaridis, Dai, and Van~Gool]{SDV21}
Christos Sakaridis, Dengxin Dai, and Luc Van~Gool.
\newblock {ACDC}: The adverse conditions dataset with correspondences for semantic driving scene understanding.
\newblock In \emph{ICCV}, 2021.

\bibitem[Touvron et~al.(2022)Touvron, Cord, and J{\'e}gou]{touvron2022deit}
Hugo Touvron, Matthieu Cord, and Herv{\'e} J{\'e}gou.
\newblock {DeiT III: Revenge of the ViT}.
\newblock In \emph{ECCV}, 2022.

\bibitem[Vu et~al.(2024)Vu, Valle, Bursuc, Kerssies, de~Geus, Dubbelman, Qian, Zhu, Chen, Tang, Wang, Vojíř, Šochman, Matas, Smith, Ferrie, Basu, Sakaridis, and Van~Gool]{vu2024bravo}
Tuan-Hung Vu, Eduardo Valle, Andrei Bursuc, Tommie Kerssies, Daan de Geus, Gijs Dubbelman, Long Qian, Bingke Zhu, Yingying Chen, Ming Tang, Jinqiao Wang, Tomáš Vojíř, Jan Šochman, Jiří Matas, Michael Smith, Frank Ferrie, Shamik Basu, Christos Sakaridis, and Luc Van~Gool.
\newblock The bravo semantic segmentation challenge results in uncv2024.
\newblock In \emph{ECCV}, 2024.

\bibitem[Wang et~al.(2023)Wang, Bao, Dong, Bjorck, Peng, Liu, Aggarwal, Mohammed, Singhal, Som, et~al.]{wang2022image}
Wenhui Wang, Hangbo Bao, Li Dong, Johan Bjorck, Zhiliang Peng, Qiang Liu, Kriti Aggarwal, Owais~Khan Mohammed, Saksham Singhal, Subhojit Som, et~al.
\newblock {Image as a Foreign Language: BEiT Pretraining for All Vision and Vision-Language Tasks}.
\newblock In \emph{CVPR}, 2023.

\bibitem[Wightman(2019)]{rw2019timm}
Ross Wightman.
\newblock Pytorch image models.
\newblock \url{https://github.com/rwightman/pytorch-image-models}, 2019.

\bibitem[Wu et~al.(2021)Wu, He, Xue, Garg, Chen, Veeraraghavan, and Barron]{wu2021train}
Yicheng Wu, Qiurui He, Tianfan Xue, Rahul Garg, Jiawen Chen, Ashok Veeraraghavan, and Jonathan~T Barron.
\newblock How to train neural networks for flare removal.
\newblock In \emph{ICCV}, 2021.

\bibitem[Zhai et~al.(2023)Zhai, Mustafa, Kolesnikov, and Beyer]{zhai2023sigmoid}
Xiaohua Zhai, Basil Mustafa, Alexander Kolesnikov, and Lucas Beyer.
\newblock {Sigmoid Loss for Language Image Pre-Training}.
\newblock In \emph{ICCV}, 2023.

\end{thebibliography}
}

\end{document}